\documentclass[letterpaper, 10 pt, conference]{ieeeconf}
\IEEEoverridecommandlockouts    % Needed for \thanks command
\let\labelindent\relax

\usepackage[table,usenames,dvipsnames]{xcolor}      % color
\usepackage{extarrows}                              % http://ctan.org/pkg/extarrows
\usepackage{enumitem}
\usepackage{wrapfig}
\usepackage{amsmath,amssymb,amsfonts,amsthm,dsfont} % math
\usepackage{algorithm,algorithmicx,listings}        % algorithms
\usepackage[noend]{algpseudocode}			              % necessary for algorithmicx
\makeatletter
\def\BState{\State\hskip-\ALG@thistlm}
\makeatother

\usepackage{graphicx}
\usepackage{tabularx}
\usepackage{subcaption}
\usepackage[font={small}]{caption}   %onehalfspacing
\usepackage[breaklinks=true, colorlinks, bookmarks=true, citecolor=Black, urlcolor=Violet,linkcolor=Black]{hyperref}

\usepackage[colorinlistoftodos,prependcaption,textwidth=1.6cm,textsize=normal]{todonotes}
\usepackage[utf8]{inputenc}
\usepackage[english]{babel}
\usepackage{hyperref}
\hypersetup{
    colorlinks=true,
    linkcolor=blue,
    filecolor=magenta,      
    urlcolor=cyan,
}

\newcommand{\prl}[1]{\mathopen{}\left(#1\right)\mathclose{}}

\DeclareMathAlphabet\mathbfcal{OMS}{cmsy}{b}{n}

\newtheorem*{assumption*}{Assumption}

\newtheorem*{problem*}{Problem}

\begin{document}

\title{Robust Fruit Counting: Combining Deep Learning, Tracking, and Structure from Motion}

% Make room for more info lines in the \author command  
\author{Xu Liu, Steven W. Chen, Shreyas Aditya, Nivedha Sivakumar, Sandeep Dcunha, \\ Chao Qu, Camillo J. Taylor, Jnaneshwar Das, and Vijay Kumar%
\thanks{The authors are with GRASP Lab, University of Pennsylvania, Philadelphia, PA 19104, USA, {\tt\small\{liuxu, chenste, ashreyas, nivedha, sdcunha, quchao, cjtaylor, djnan, kumar\}@seas.upenn.edu}.}%
}
%Use only for final RAL version. 

%\thispagestyle{empty}
%\pagestyle{empty}
% Paper headers 
\markboth{IEEE Robotics and Automation Letters. In preparation}
{Xu \MakeLowercase{\textit{et al.}}: Robust Fruit Counting}  

\maketitle
%%%%%%%%%%%%%%%%%%%%%%%%%%%%%%%%%%%%%%%%%%%%%%%%%%%%%%%%%%%%%%%%%%%%%%%%%%%%%%%%
\begin{abstract}
We present a novel fruit counting pipeline that combines deep segmentation, frame to frame tracking, and 3D localization to accurately count visible fruits across a sequence of images. Our pipeline works on image streams from a monocular camera, both in natural light, as well as with controlled illumination at night. We first train a Fully Convolutional Network (FCN) and segment video frame images into fruit and non-fruit pixels. We then track fruits across frames using the Hungarian Algorithm where the objective cost is determined from a Kalman Filter corrected Kanade-Lucas-Tomasi (KLT) Tracker. In order to correct the estimated count from tracking process, we combine tracking results with a Structure from Motion (SfM) algorithm to calculate relative 3D locations and size estimates to reject outliers and double counted fruit tracks. We evaluate our algorithm by comparing with ground-truth human-annotated visual counts. Our results demonstrate that our pipeline is able to accurately and reliably count fruits across image sequences, and the correction step can significantly improve the counting accuracy and robustness. Although discussed in the context of fruit counting, our work can extend to detection, tracking, and counting of a variety of other stationary features of interest such as leaf-spots, wilt, and blossom.\\
% Key Words: Deep Learning in Robotics and Automation, Agriculture Automation, Visual Tracking, Object Segmentation, Detection, Categorization

\end{abstract}

\section{Introduction}

\label{sec:intro}
\begin{figure}[t!]
\vspace*{0.08in}
\centering
\includegraphics[width=\columnwidth]{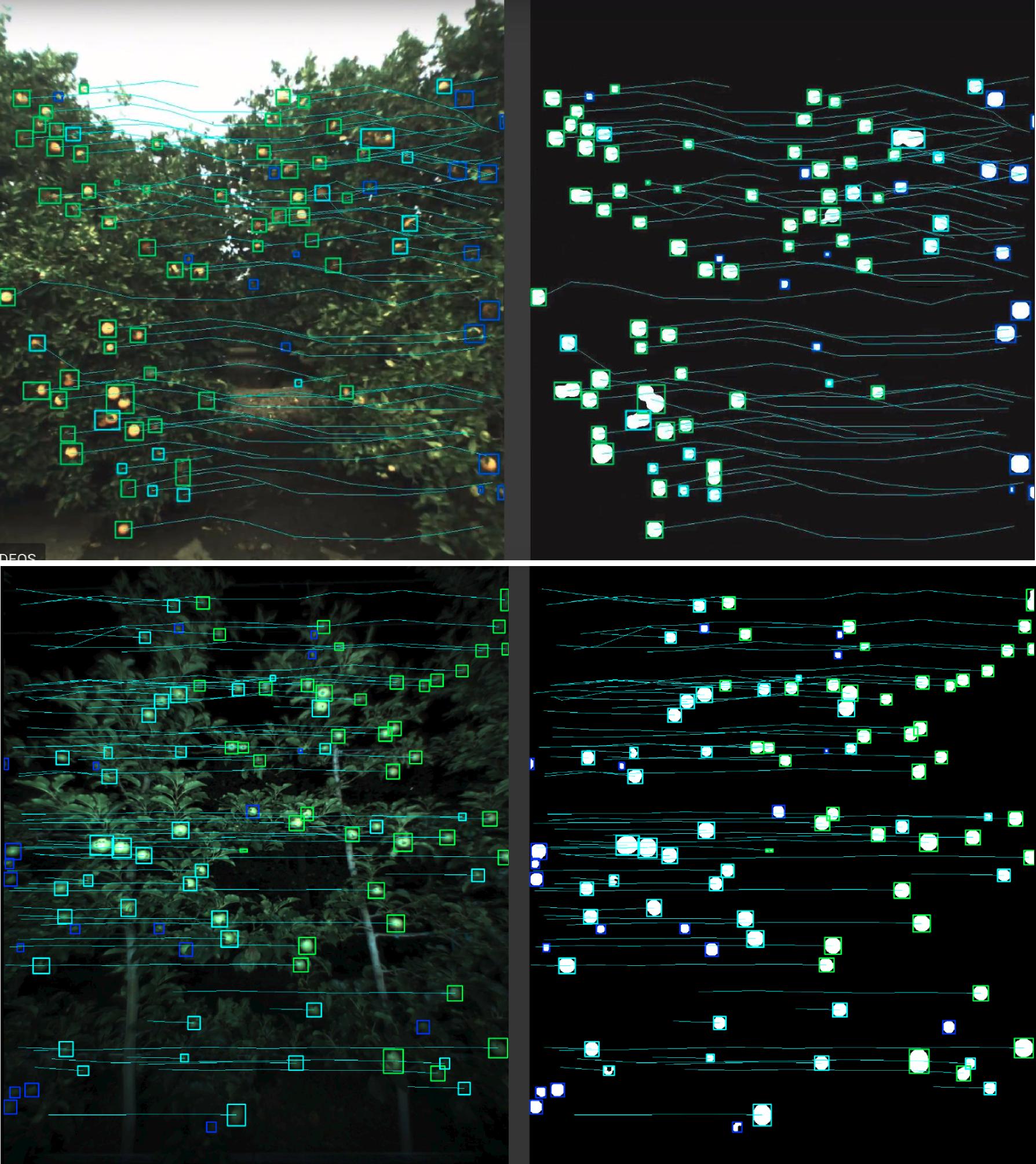}
\caption{\textit{Fruit tracking by detection:}
The top two panels show oranges during the day, and the bottom two panels show green apples at night. The count obtained by our detection and tracking algorithm is corrected by using 3D locations and relative size estimates of the fruits to remove outliers and double counted fruit tracks.}
\label{fig:front-page}
\vspace*{-0.25in}
\end{figure}

The ability to obtain fruit counts from videos allows growers to better optimize management and harvest decisions such as labor allocation, storage, packaging, and harvest scheduling. Despite recent progress in using deep learning to improve fruit detection from static images, counting from videos still remains a challenge due to outliers originating from fruit tracking and localization errors~\cite{das2015devices,chen2017,roy2016}. The problem of getting the fruit counts from a video can be addressed using the framework \textit{tracking by detection}~\cite{leibe2007coupled} consisting of two phases: fruit detection, followed by tracking and counting (Fig.~\ref{fig:front-page}). 

%In this paper, we present a fruit counting pipeline that combines fruit segmentation using Fully Convolutional Networks (FCN), tracking using the Hungarian Algorithm with the Kanade-Lucas-Tomasi (KLT) Tracker and a Kalman Filter, and 3D localization using a Structure from Motion (SfM) algorithm. Our algorithm works on image streams from a monocular camera, and has been tested on both daytime oranges and nighttime green apple datasets. 

%This pipeline results in better counts through removal of outliers that originate from the tracking process, or from extraneous detections due to fruits on the ground and in other rows.

%We found that compared to using SIFT features from the whole image, using SIFT features from segmented fruits only results in better camera pose estimation, consequently, improving 3-D localization and relative size estimates. 

\begin{figure*}[t!]
\centering
\includegraphics[width=6.7in]{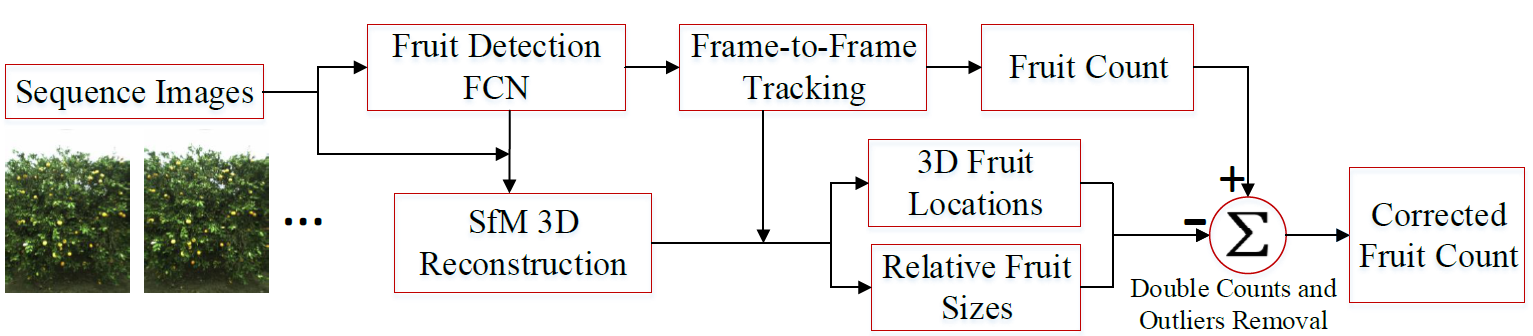}
\caption{{\small Our fruit counting pipeline consists of detection, tracking, and 3D localization stages.} The segmentation stage uses an FCN to segment image into fruit and non-fruit pixels. The tracking stage then tracks the fruits based on the FCN outputs. The 3D localization stage uses SfM reconstruction to estimate fruit 3D locations and size. Finally, based on these localization and size estimation results, we correct the total fruit count from tracking stage, and obtain the corrected fruit count.}
\label{fig:pipeline}
\vspace*{-0.2in}
\end{figure*}

% %1. fruit counting background, why is important?
% 2. Segmentation, why FCN is good choice
% 3. Tracking + Kalman filter: obtain robust frame-frame correspondence, avoid double counting
% 4. localization and size estimation: help us a lot in calibrating fruit count (avoid double tracks, avoid outliers), and have many potential applications
%  A detailed understanding of the yield variation across an orchard allows growers to make better decisions for labor allocation, storage, packaging, and harvesting. 
%The availability of camera sensors ranging from customized sensor suites or robots such as unmanned aerial vehicles (UAVs) to ordinary smartphones with cameras, means that our pipeline can benefit growers of all sizes and financial resources around the world. The ideal use case of our pipeline is a farmer either deploying an automated robotic system, or simply walking with a smartphone, to take a video sequence of the trees in her orchard. The algorithm should then process this video sequence and return the total number of fruit in the video.
%The problem of getting the fruit counts from a video can be addressed using the framework \textit{tracking by detection}~\cite{leibe2007coupled}. This framework consists of two phases: fruit detection, followed by tracking and counting. The first detection phase involves detecting candidate regions in each individual image frame. Next, the second tracking phase connects the detections by tracking them across frames (Fig.~\ref{fig:front-page}). 
Applying this framework to the fruit counting problem is challenging for a variety of reasons. First, the algorithm needs to generalize across different fruit types and environments that can vary greatly based on illumination conditions, tree shapes, orchard arrangements, etc. While, most previous algorithms use hand-engineered features for each specific scenario to detect the fruit~\cite{pothentexture}, recent works exploit deep learning algorithms that perform well across a variety of conditions~\cite{chen2017,bargoti2017deep,sa2016deepfruits}. Second, after obtaining detections for each image frame, these individual detections must be tracked across the frames in order to prevent over-counting of the same fruit. This step is also challenging in unstructured environments, as double-counting of fruit can still result from occlusion or illumination variation. 

%Finally, the algorithm should only count fruit from the target row, while ignoring visible fruit on the ground or in a further row.
%is challenging because the lighting condition?????????????????????
%This paper introduces a pipeline that expands upon our previous work on tracking and counting in 2-D with a hand-designed detection algorithm~\cite{das2015devices}, and deep segmentation models for accurate detection~\cite{chen2017}. This improvement is achieved by integrating the approaches in both methods, as well introducing a 3D localization step that eliminates double counted fruit and outliers in order to improve the fruit count. 
%with a few additions that further improve performance. ?????????

This paper introduces a pipeline that expands upon our previous works on tracking and counting in 2D with a hand-designed detection algorithm~\cite{das2015devices}, and accurate detection using deep segmentation models~\cite{chen2017}. The first step of our pipeline uses a sample set of human-generated ground-truth labels to train a Fully Convolutional Network (FCN) that segments individual frame images into fruit and non-fruit pixels. The next step uses the Hungarian Algorithm with an objective cost computed from using a Kalman Filter corrected KLT tracker. The third step localizes the fruit in 3D using a Structure from Motion (SfM) algorithm that tracks Scale-Invariant Feature Transform (SIFT) features that are located near the fruit regions. This last stage corrects the fruit count by eliminating double counted fruits and potential outliers in terms of size and location. In addition, these size estimates are useful in estimating yield and analyzing crop growth. We evaluate our algorithm by comparing with human annotated ground-truth visual count on orange and apple data and report accuracy both before and after correction. These results demonstrate accuracy and consistency of our algorithm in counting fruits across image sequences.

The main \textbf{contributions} of our paper are: (1) to the best of our knowledge, the first application of a deep learning segmentation network into a pipeline that counts fruits across an image sequence obtained from a monocular camera; and (2) the introduction of a count correction based on 3D localization and size distribution estimates that significantly improves the accuracy and error standard deviation of the count estimates by rejecting false positives. A video of our algorithm can be found at: \url{https://youtu.be/5pSemkmVMRc} or \url{http://label.ag/iros18.mp4}.

\section{Related Work}
Deep learning techniques have recently replaced traditional hand-engineered computer vision techniques as the state of the art for fruit detection, segmentation and counting in a single image. The Faster Regions with Convolutional Neural Networks (Faster R-CNN) architecture has been used for detection of mangoes, almonds, apples, and a variety of other fruits~\cite{bargoti2017image,bargoti2017deep, sa2016deepfruits}. Chen et al. use the Fully Convolutional Network (FCN) to segment the image into candidate regions and CNN to count the fruit within each region~\cite{chen2017}. Rahnemoonfar and Sheppard train an Inception style architecture to directly count the number of tomatoes in an image, demonstrating that in some scenarios these deep networks can even be trained using synthetic data~\cite{rahnemoonfar2017deep}. Barth et al. also generate synthetic data to train deep neural networks to segment pepper images~\cite{barth2018data}. Our work differs from these previous works by expanding the counting problem from a single image to an image sequence. This extension is more challenging since it requires tracking of the fruits across image frames.

Most approaches that track fruits across image frames use some combination of Structure from Motion (SfM), Hungarian algorithm, Optical Flow, and Kalman Filters to provide the corresponding assignments of fruits across frames. Roy and Isler~\cite{roy2016surveying} use apple contours in an SfM pipeline to generate dense matches and reconstruct the apples in 3D. They use this 3D reconstruction to count and measure the apple diameters. Das et al. use a Support Vector Machine (SVM) to detect fruits, and use optical flow to associate the fruits in between frames~\cite{das2015devices}. Wang et al. use stereo cameras to count red and green apples by taking images at night in order to control the illumination to exploit specular reflection features~\cite{wang2013automated}. They get the 3D locations of each apple in order to merge multiple detections of the same fruit, and use these 3D locations to count the total number of fruit. 

These previous tracking approaches differ from our work because they use traditional computer vision techniques to segment the fruit from the images. As a result, they must rely on controlled illumination of the orchards, which previous work has noted as a limitation~\cite{das2015devices}. Thus, our primary contribution is the integration of deep learning methods for fruit segmentation with SfM techniques for fruit tracking, resulting in a generalizable and robust monocular pipeline that removes double-counted fruits by exploiting 3D locations and relative size distribution estimation.

\section{Problem Formulation}
Consider a sequence of $n$ images $\mathbfcal{I} = (\mathcal{I}_{k})_{k=1}^{n}$ of trees containing fruit of the same type. The image sequence $\mathbfcal{I}$ has a state $\mathbf{z} \in \mathbb{N}$ representing the count of visible fruit in this image sequence $\mathbfcal{I}$.

\textbf{Visible Fruit Counting:} Given a set of $m$ image sequences $\{\mathbfcal{I}^{(j)}\}_{j=1}^{m}$, our goal is to find the mapping $f \colon \mathbfcal{I}^{(j)} \rightarrow \mathbf{z}^{(j)} \in \mathbb{N}$ to produce  visual count from the image sequence set. The discrepancy between this estimate  $\mathbf{z}^{(j)}$ and the true visual count $\hat{\mathbf{z}}^{(j)}$ is captured through the L1 loss, defining the metric used for the evaluation of our algorithm. 
\begin{equation}
%\sum_{j=1}^{m}|f(\mathbf{\mathcal{I}}^{(j)})-\mathbf{z}^{(j)}|
\sum_{j=1}^{m}|f(\mathbf{\mathcal{I}}^{(j)})-\hat{\mathbf{z}}^{(j)}|
%\vspace*{-0.1in}
\end{equation}

\section{Proposed Approach}
% Xu: Our Proposed Approach
% \begin{figure*}[htpb]
% \centering
% \includegraphics[width=7in]{imgs/3D-fruit-counting.png}
% \caption{{\small The fruit counting pipeline}}
% \label{fig:fruit-counting}
% \end{figure*}
We break the overall \textit{visible fruit counting} problem into three sub-problems: (1) Segmentation of candidate fruit regions, (2) Tracking of regions between frames, and (3) 3D reconstruction in order to detect false positives generated during the segmentation and tracking steps. As depicted in Fig.~\ref{fig:pipeline}, our approach addresses each subproblem and combines the individual outputs together in order to generate the estimated fruit count of the entire image sequence. Use in low-resource setting is a potential future scope of our pipeline, where the imagery may only be obtained from off-the-shelf cameras. As a result, we limit our sensor inputs only to RGB images generated by a monocular camera. 

\begin{figure}[t!]
\centering
\includegraphics[width=3.0in]{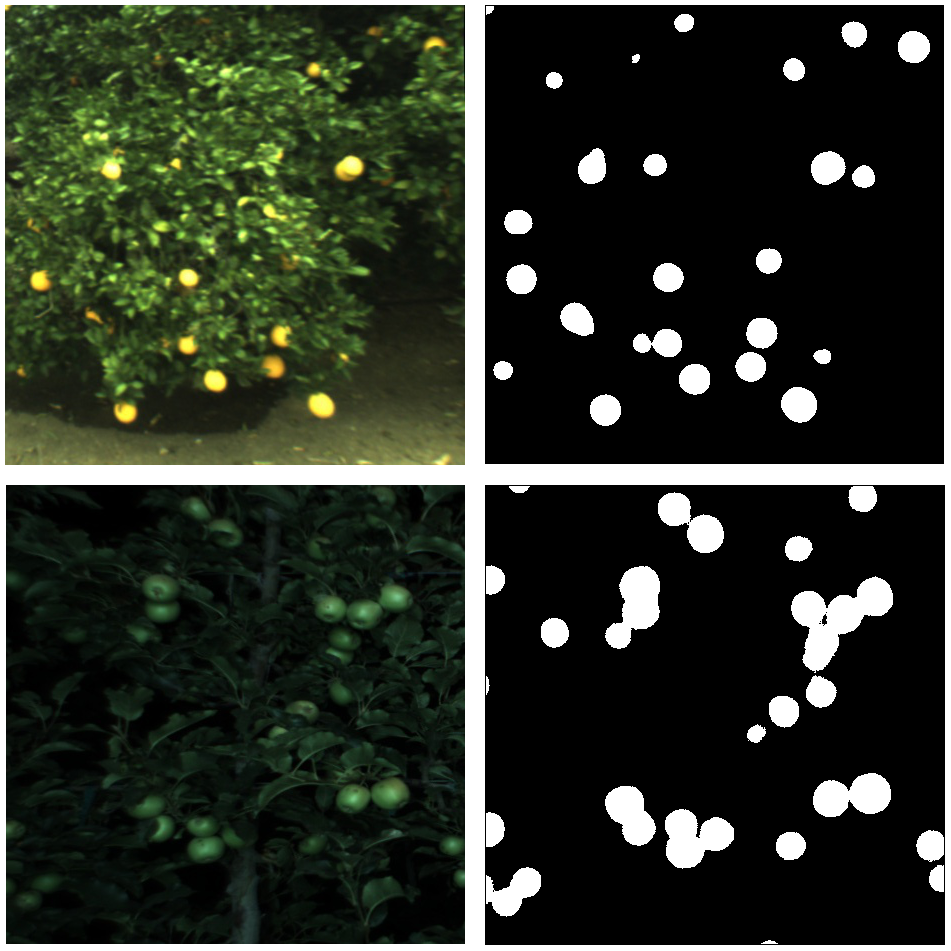}
\caption{\textit{Segmentation Results for daytime orange (top), and nighttime green apples (bottom).} The FCN network is able to detect most fruits, including highly occluded ones, in both datasets.}
\label{fig:seg_results}
\vspace*{-0.25in}
\end{figure}

\subsection{Segmentation with Deep Learning}

Many recent fruit detection and segmentation works have shown state-of-the-art performance with various deep learning algorithms~\cite{bargoti2017deep,sa2016deepfruits,li2016deep}. We expand on our previous work~\cite{chen2017} by employing the same Fully Convolutional Network (FCN) architecture~\cite{long_shelhamer_fcn} used in segmenting candidate fruit regions.

\textbf{Obtaining ground truth}: Our previous work has noted that after utilizing standard data augmentation techniques~\cite{krizhevsky2012imagenet}, training the FCN for fruit segmentation using only a few labeled images as data, achieves reasonable performance due to the high similarity in fruit appearances. However, obtaining large amounts of labeled data is still important to increase generalization ability and to perform evaluation. 

\begin{figure}[t!]
\vspace*{0.01in}
\centering
\includegraphics[width=3.253in]{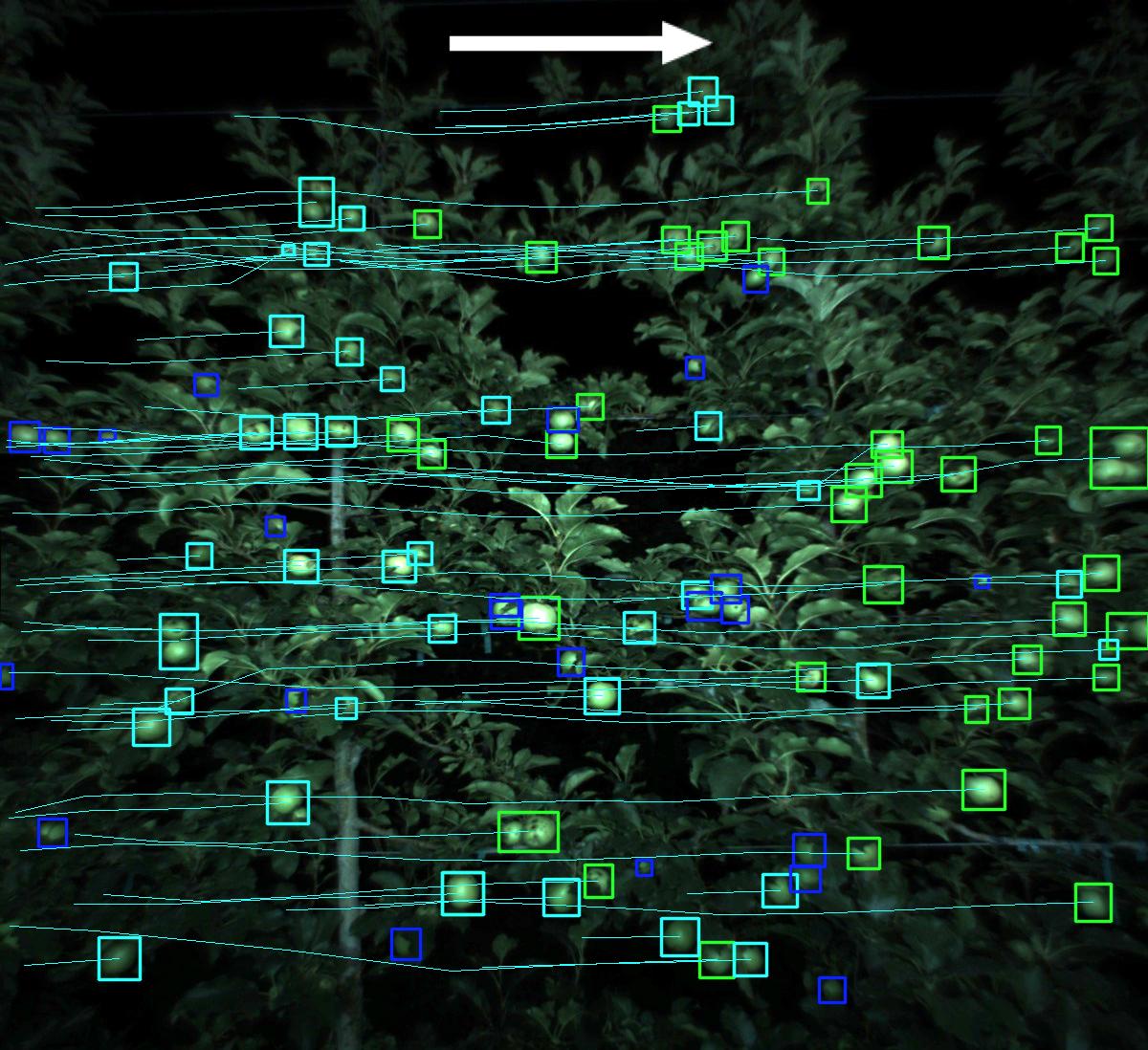}
\caption{\textit{Tracking result:  (Lines: fruit trajectories. Boxes: green: counted fruits, cyan: tracked fruits, blue: new detections). The arrow shows the opposite direction of camera motion.} The trajectories show that the tracking process is stable. Most fruits are recorded as counted fruits before disappearing, and most false positives are either \textit{new detections} or \textit{tracked fruits}. Since we only count \textit {counted fruits}, such false positives have no negative influence in final count.}
\label{fig:tracking_result}
\vspace*{-0.25in}
\end{figure}

In our previous work, we developed a web-based labeling framework called \url{label.ag} to quickly collect and store ground truth labels from a team of human labelers. Instead of pixel-wise label masks, the labels are represented in the vector based Scalable Vector Graphics (SVG) format that stores locations and radii of circles. This SVG representation is a more flexible representation since the circles can be used to generate other label representations such as label masks, and fruit counts.

\textbf{Fruit segmentation network}: 
Our neural network architecture is an FCN that segments the image into fruit pixels and non-fruit pixels. The FCN differs from classical convolutional networks because it does not utilize any fully connected layers. Utilizing only convolutional and deconvolutional layers allows the network to perform segmentation tasks instead of classification or regression. 

The FCN takes in an image and outputs a score map of size \textit{$h \times w \times n$}, where $\textit{h}$ and $\textit{w }$ are the height and width of the image, and n is the number of classes. Every pixel has its own score of class participation. For example, the score for pixel $(i,j)$ belonging to class  $\textit{k }$ is represented by $x_{ijk}$  element in the score map. Our FCN has the same structure as the original FCN~\cite{long_shelhamer_fcn}, which is based on the VGG Network architecture used in image classification~\cite{simonyan2014very}. The main difference is that our network has 2 classes (fruit and not-fruit) rather than the original 21 classes. Using the same architecture allows for quick training as we can initialize weights as the VGG network weights. An example of our segmentation results is illustrated in Fig.~\ref{fig:seg_results}.

\subsection{Tracking between frames}
The fruit candidate region output from our deep learning fruit segmentation stage is used as the input into the tracking stage. This part of our pipeline is used to obtain association across the image frames, and is a necessary step to obtain fruit counts for the entire image sequence as opposed to an individual image.

The tracking task is essentially a \textit{Multiple Hypothesis Tracking} (MHT) problem~\cite{reid1979algorithm}, which is extensively discussed in a variety of contexts, including fruit tracking~\cite{das2015devices}. The input to our MHT problem is a sequence of binary image masks representing pixel-wise classification of fruit and non-fruits, and the output are fruit objects parameterized by spatial location $u$ and $v$ in pixel space and frame $k$. 

The first step is to convert pixels into fruit objects. Since the pixel-wise outputs of the FCN will have fruit pixels that are naturally clustered together; this step is simplified as we can perform contour analysis. The region inside every contour represents our initial guess for the fruit.

We now need to assign the fruits from image $\mathcal{I}_{k}$ to the next image $\mathcal{I}_{k+1}$. We can formulate our problem as a graph problem. Supposing we have a complete bipartite graph $\mathcal{G} = (\mathcal{S}, \mathcal{T}; E)$ where the vertices $s \in \mathcal{S}$ represent the candidate fruit regions in image $\mathcal{I}_{k}$, and the vertices $t \in \mathcal{T}$ are the candidate fruit regions in image $\mathcal{I}_{k+1}$. Since the graph is complete, there is an edge $(s,t) \in E~\forall s \in \mathcal{S}, t \in \mathcal{T}$. Each edge $(s,t)$ has an associated cost $c(s,t)$ which can be arbitrarily chosen. Finding the perfect matching with minimum cost of graph $\mathcal{G}$ will yield the corresponding tracks across the frames that minimizes our chosen cost function. The Hungarian Algorithm~\cite{munkres1957algorithms} is the standard algorithm used to solve this problem.

In our context, the interesting choice is in designing the cost function $c(s,t)$ between candidate fruit regions for $\mathcal{I}_{k}$ and $\mathcal{I}_{k+1}$. Suppose vertex $s$ corresponds to fruit $i$ in frame $k$, and vertex $t$ to fruit $j$ in frame $k+1$. Let $p_{i,k} = [u_{i,k}, v_{i,k}]^{T}$ represent the row $u_{i,k}$ and column $v_{i,k}$ position of fruit region $i$ in the image coordinate space of image $\mathcal{I}_{k}$. One simple cost function would be the Euclidean distance between the centers $c(s,t)=||p_{i,k}-p_{j,k+1}||_{2}^{2}$. However, since the fruit objects are dense and the camera is moving, the fruit region in image $\mathcal{I}_{k}$ and its corresponding closest fruit region in image $\mathcal{I}_{k+1}$ may represent different fruits. Therefore, this approach will frequently yield wrong matches.

Instead, we define a different cost function based on a Kalman filter corrected optical flow estimate. The optical flow tracker we use is the Kanade-Lucas-Tomasi (KLT) feature tracker~\cite{Lucas:1981:IIR:1623264.1623280}. The KLT tracker takes in images $(\mathcal{I}_{k}, \mathcal{I}_{k+1})$ along with the desired fruit center to track $p_{i,k}$ in image $\mathcal{I}_{k}$, and determines the optical flow $d_{i,k} = [d^{(u)}_{i,k}, d^{(v)}_{i,k}]^{T}$ at point $p_{i,k}$. If we just use the optical flow with no Kalman Filter correction, our cost function would be ${c(s,t) = ||\prl{p_{i,k}+d_{i,k}} - p_{j,k+1}||_{2}^{2}}$. Notice that the optical flow step only uses the FCN segmentation in image $\mathcal{I}_{k}$, and the new FCN segmentation in image $\mathcal{I}_{k+1}$ is only introduced in the Hungarian Algorithm cost function.

 Just using the KLT algorithm will sometimes yield noisy flow where different fruits can be predicted to move in very different directions. This is not desirable when the fruits are stationary and the camera is moving sideways, which is the most common case for fruit counting videos. As a result, the directions of optical flow should be similar between fruits. 
 
We address this problem using the following two techniques. First, we use the Kalman Filter~\cite{kalman1960new}, also known as the Linear Quadratic Estimator (LQE), to correct the optical flow estimates. Second, we use the velocity of every fruit as an initial guess of its optical flow for the KLT tracker. This velocity is part of the state vector in the Kalman Filter, and is designed to account for the motion of the camera.

Define the expanded $4 \times 1$ state vector $x_{i,k}$ for fruit center $i$ at frame $k$ as:
\begin{equation*}
x_{i,k} = \begin{bmatrix}
           p_{i,k} \\
           \dot{p}_{i,k} \\
         \end{bmatrix}
         = \begin{bmatrix}
           u_{i,k} \\
           v_{i,k} \\
           \dot{u}_{i,k} \\
           \dot{v}_{i,k} \\
         \end{bmatrix}, 
\end{equation*}
where we now include $\dot{u}_{i,k}$ the pixel row velocity, and $\dot{v}_{i,k}$ the pixel column velocity. 

We use the following discrete-time time-invariant linear system model
\begin{equation}
\begin{aligned}
x_{i,k+1} &= \mathbf{A}x_{i,k} + \omega \\
z_{i,k+1} &= \mathbf{H}x_{i,k} + \nu
\end{aligned}
\label{eqn:process_model}
\end{equation}
where $\mathbf{A}$ is the state transition matrix, $\mathbf{H}$ is the observation matrix, $\omega$ the process noise, and $\nu$ the measurement noise. Both $\omega$ and $\nu$ are random variables assumed to be drawn from a Gaussian zero-mean distribution. Specifically, the parameters of this model are defined as 
{\small \begin{equation*}
\begin{aligned}
\mathbf{A} = \begin{bmatrix}
    1 & 0 & 1 & 0 \\
    0 & 1 & 0 & 1 \\
    0 & 0 & 1 & 0 \\
    0 & 0 & 0 & 1 \\
\end{bmatrix}&, 
\hspace{0.1in}
\mathbf{H} = \begin{bmatrix}
    1 & 0 & 0 & 0 \\
    0 & 1 & 0 & 0\\
    0 & 0 & 1 & 0 \\
    0 & 0 & 0 & 1 \\
     
\end{bmatrix} \\
\omega & \sim \mathcal{N}(0, \mathbf{Q})\\
\nu & \sim \mathcal{N}(0, \mathbf{R})
\end{aligned}
\end{equation*}}
where $\mathbf{Q}$ and $\mathbf{R}$ are user-defined $4 \times 4$ covariance matrices. Since the major movement of fruits is along the $u$ (row) direction, we set larger noise values for variables in the $u$ direction. Our optical flow estimates constitute our noisy observation variable as follows
\begin{equation*}
z_{i,k+1} = \begin{bmatrix}
    u_{i,k} + d^{(u)}_{i,k} \\
    v_{i,k} + d^{(v)}_{i,k} \\
    \frac{1}{m}\sum_{l=1}^{m}d^{(u)}_{l,k-1} \\
    \frac{1}{m}\sum_{l=1}^{m}d^{(v)}_{l,k-1} \\
\end{bmatrix}, \\
\end{equation*}
where $m$ is the total number of candidate fruit regions in image $\mathcal{I}_{k-1}$. In other words, to measure the predicted state at frame $k+1$ we use the optical flow predicted locations from the current frame $k$ as the measured position variables, and the average of the optical flow in the previous frame $k-1$ as the measured velocity variable. This averaging helps to smooth out the noisy flow problem we highlighted before. 

Thus, given a state $x_{i,k}$, using the process model in Eqn~\eqref{eqn:process_model}, we will get an \textit{a priori} estimate $\hat{x}^{-}_{i,k+1}$ for frame $k+1$ given knowledge of the process prior to step $k+1$. Using the measurement $z_{i,k+1}$, we perform the standard Kalman Filter prediction and update steps detailed in~\cite{kalman1960new} to compute the \textit{a posteriori} state estimate $\hat{x}_{i,k+1}$, and then extract the position variables to get estimate of $\hat{p}_{i,k+1} = [\hat{u}_{i,k+1}, \hat{v}_{i,k+1}]^{T}$. The cost function is then $c(s,t) = ||\hat{p}_{i,k+1} - p_{j,k+1}||_{2}^{2}$. 

The above cost function only takes into account the distance between the predicted center and the next center. We make a few modifications to also take into account the fruit size measured by its bounding box and the overlap between bounding boxes. When we determine the fruit correspondences, we care about the center distance relative to the size of the bounding boxes. In addition, we would like the bounding boxes to overlap, as this indicates that they are more likely to be a good match.

\begin{figure*}[t!]
\centering
\includegraphics[width=6.5in]{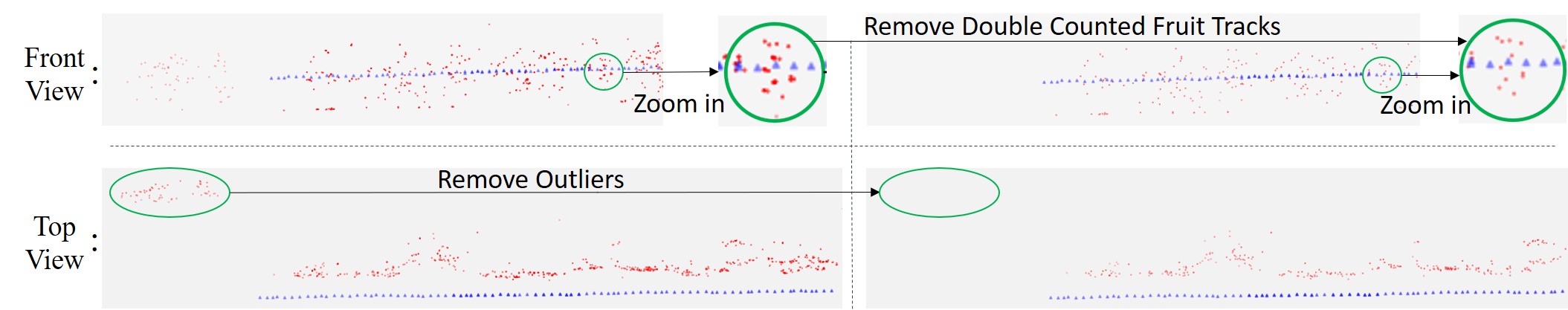}
\caption{\textit{Fruits 3D localization result before (left) and after (right) rejecting outliers and double counted tracks.} The red dots are estimated fruit locations and the blue dots are camera centers for every frame image. Note how the fruits from the tree behind the target row are removed due to the knowledge of 3D fruit locations. Double counted fruit tracks are very close in 3D (ideally, they should totally overlap), but they are projected from different sequence images, thus being hard to detect in a 2D image plane.}
\label{fig:outlier_rejection}
\vspace*{-0.2in}
\end{figure*}

Suppose that $a_{i,k}$ and $a_{j,k+1}$ are the areas of the bounding boxes for fruit $i$ in image $\mathcal{I}_{k}$ and fruit $j$ in image $\mathcal{I}_{k+1}$. In addition, let $\lambda_{k+1}(i,j)$ be the proportion overlap of the bounding box of fruit $i$ after it's center has been estimated to move to $\hat{p}_{i,k+1}$ from the Kalman Filter corrected Optical Flow step, with the bounding box of fruit $j$ in $\mathcal{I}_{k+1}$. Our final cost function is:
\begin{equation*}
c(s,t) = \frac{||\hat{p}_{i,k+1} - p_{j,k+1}||_{2}^{2}}{a_{i,k} + a_{j,k+1}} + \prl{1 - \lambda_{k+1}(i,j)}
\end{equation*}

After performing the Hungarian Algorithm assignment, we re-initialize the Kalman Filter model with all the new candidate regions in the next frame where for fruit $i$ in $\mathcal{I}_{k+1}$:
\begin{equation*}
x_{i,k+1} = \begin{bmatrix}
    u_{i,k+1}\\
    v_{i,k+1} \\
    \frac{1}{m}\sum_{l=1}^{m}d^{(u)}_{l,k} \\
    \frac{1}{m}\sum_{l=1}^{m}d^{(v)}_{l,k} \\
\end{bmatrix} \\
\end{equation*}
and the covariance matrices are re-initialized accordingly.

The final step is to determine which fruit tracks to count. Rather than counting all fruit tracks, we instead select a subset of the fruit tracks based on their age, i.e., the total number of frames that the fruit has been tracked. Only fruit tracks with an age above the predefined threshold will be counted, and are counted when we lose track of them. In our implementation, to achieve a good trade-off between false positive and false negative tracks, this threshold is defined to be one third of the number of frames which have overlap. Differentiating fruit tracks according to their age can make our counting more robust.

% The threshold to differentiate between those three is determined by the number of overlapping frames of the video. For example, if there are $2\textit{m}$ frames have overlap, newly detected fruits only appear in the current frame(age=1); tracked fruits appear in at least 2 frames and at most $\textit{m}$ frames (2$\leq$age$<\textit{m}$). Counted fruits appear in at least  $\textit{m}$ frames ($\textit{m}\leq$age), and are added to the total count of fruits immediately when we lose track of them. The reason of differentiating between tracked fruits and counted fruits is to reduce false positive tracks, making the tracking algorithm more robust.

\subsection{3D Fruit Localization and Relative Size Estimation}

\begin{figure}[t!]
\centering
\includegraphics[width=255px]{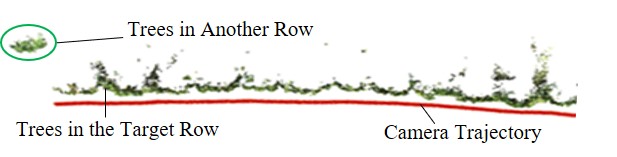}
\caption{\textit{Top view of 3D Reconstruction of Trees.} We can see the distribution of tree canopy is consistent with the distribution of fruits in Fig.~\ref{fig:outlier_rejection}, and combine the two we can intuitively see how fruits from another row are removed.}
\label{fig:3D reconstruction}
\vspace*{-0.25in}
\end{figure}

The previous detection and tracking steps will provide an initial count of the number of fruits in an image sequence. However, as depicted in the reconstruction result in Fig.~\ref{fig:3D reconstruction} and the fruit localization result in Fig.~\ref{fig:outlier_rejection}, this count is still susceptible to fruits that are double counted, and fruits from trees in other rows behind the one being counted, but still visible to  the camera. To account for these pitfalls, this count can thus be further corrected by localizing the fruit in 3D and estimating the relative sizes. 

We obtain our 3D reconstruction using an incremental Structure from Motion (SfM) algorithm. SfM matches features across image frames to estimate camera poses. A common feature to track are Scale Invariant Feature Transform (SIFT) features~\cite{lowe2004distinctive}. Instead of tracking all SIFT features, we only track SIFT features located close to our detected fruit regions. The intuition for this modification is that tree environments are highly textured, and as a result the best features to track are those close to the objects of interest. We found that this reduction of features not only decreases computation time, but also improves the overall quality of the 3D reconstruction.

In order to only detect SIFT features near fruit, we first identify the fruit bounding boxes and add a margin of 25 pixels to each side. We then use a $5 \times 5$ window to apply an average blur around the edge. This blur helps to avoid false positives from an artificially high contrast. Next, we set the background pixels to be the average pixel intensity value of the lower third region of all the images in the dataset. This constant color choice removes the features that are not close to fruit centers. In addition, it approximates the color of the leaves which reduces false positives by approximating the color of leaves. Compared with using all visual features, only using features near fruits significantly reduces the total number of matched features, provides a even spread of matched features throughout the image, and maintains the number of features lying on fruits. We found that this not only saves the computation power, but also improves the quality of our reconstruction. 

\begin{figure*}[t!]
\begin{center} 
 \setlength\extrarowheight{5pt}
 \resizebox{\textwidth}{!}
 {\begin{tabular}{||c | c | c | c | c||} 
 \hline
 Measure & Oranges (Uncorrected) & Oranges (Corrected) &  Apples (Uncorrected) &  Apples (Corrected)\\ [0.6ex] 
 \hline\hline
 Count / \textit{Ground Truth} & 4049 / \textit{3456} & \textbf{3449 / \textit{3456}} & 8622 / \textit{7949} & \textbf{8215 / \textit{7949}}\\
 \hline\hline
 L1 Loss & 593 & \textbf{203} & 673 & \textbf{322}\\
  \hline\hline
 Error Mean & 17.2$\%$ & \textbf{-0.2}$\%$& 8.5$\%$ & \textbf{3.3}$\%$\\
  \hline\hline
 Error Std Dev  & 26.3$\%$ & \textbf{7.8}$\%$ & 4.7$\%$ & \textbf{4.1}$\%$\\
    \hline\hline
\end{tabular}}
\end{center}
\caption{\textit{Error mean and standard deviation of our algorithm on the orange and apple datasets}: The correction from the 3D fruit localization significantly improves the algorithm performance by reducing both the error mean and standard deviation. The correction is effective especially for the orange data set, in which the environment is not as carefully controlled, and the fruits have more variation in depth. }
\label{fig:oranges_count_accuracy}
\vspace*{-0.2in}
\end{figure*}

However, one limitation with this approach is that occasionally, there are not enough features around fruits. The lack of features around fruits can occur due to few number of fruits appearing in some frames and the smoothness of the areas around some fruits. When such scenarios occur, the 3D reconstruction based on features around fruits will be unstable. To increase robustness, if enough features are not detected near fruits, we allow the SfM algorithm to use all visible features.

After the 3D SfM reconstruction, we will have the 3D locations of each SIFT feature. To compute the 3D location of the fruit centers, we average the 3D locations of each feature associated with that fruit. Once these fruit centers have been located, we can compute the relative depth by projecting the 3D fruit center from the global frame to the local camera frame. The relative sizes can then be estimated by multiplying the fruit area in image pixels with the square of the relative depth.

% \begin{figure}[htpb]
% \centering
% \includegraphics[width=7in]{imgs/remove_outliers.jpg}
% \caption{\textit{Top view of fruits and camera poses, before (top) and after (bottom) outlier rejection. Note how the fruits from the tree behind the target row is removed due to the knowledge of 3D fruit locations.\todoJD{Fig.~\ref{fig:outlier_rejection} needs annotations for what the green region is, and a divider between the outliers and no-outliers cases.}}}
% \label{fig:outlier_rejection}
% \vspace*{-0.2in}
% \end{figure}

The 3D fruit locations and relative sizes can now be used to correct the initial fruit count. First we can remove double counted fruits. Due to occlusion and illumination variation, not every fruit will be tracked in all the frames they appear in. As a result, some fruits that are initially tracked, then lost, and then detected and tracked again at a later frame, will be counted as different fruits and appear twice in our initial count. After localizing these fruit tracks in 3D, we can easily detect these double counted fruits since they will be close in 3D space. 

In addition, we can remove fruit outliers in terms of size. These outliers can be detected by comparing the relative size distribution, where both too large and too small fruits can be regarded as false positives. For example, in some datasets, signs can sometimes be classified as a fruit. This step will detect and eliminate these false positives since the estimated size will be large. The size-based outliers rejection is also depicted in Fig.~\ref{fig:rel_size_orange}, fruits with sizes smaller than lower threshold (blue line) and upper threshold (red line) are all regarded as outliers.

We can also reject fruits that do not lie in the row we are counting. Such a scenario is illustrated in Fig.~\ref{fig:outlier_rejection}, when a further row of trees is visible due to a gap in the tree row we are counting. Since fruits' 3D locations will be estimated, we can then project them from world coordinate to camera frame, and set a threshold to the z coordinate which represents the depth.

\section{Results and Analysis}

\begin{figure*}[t!]
\centering
\includegraphics[width=7in]{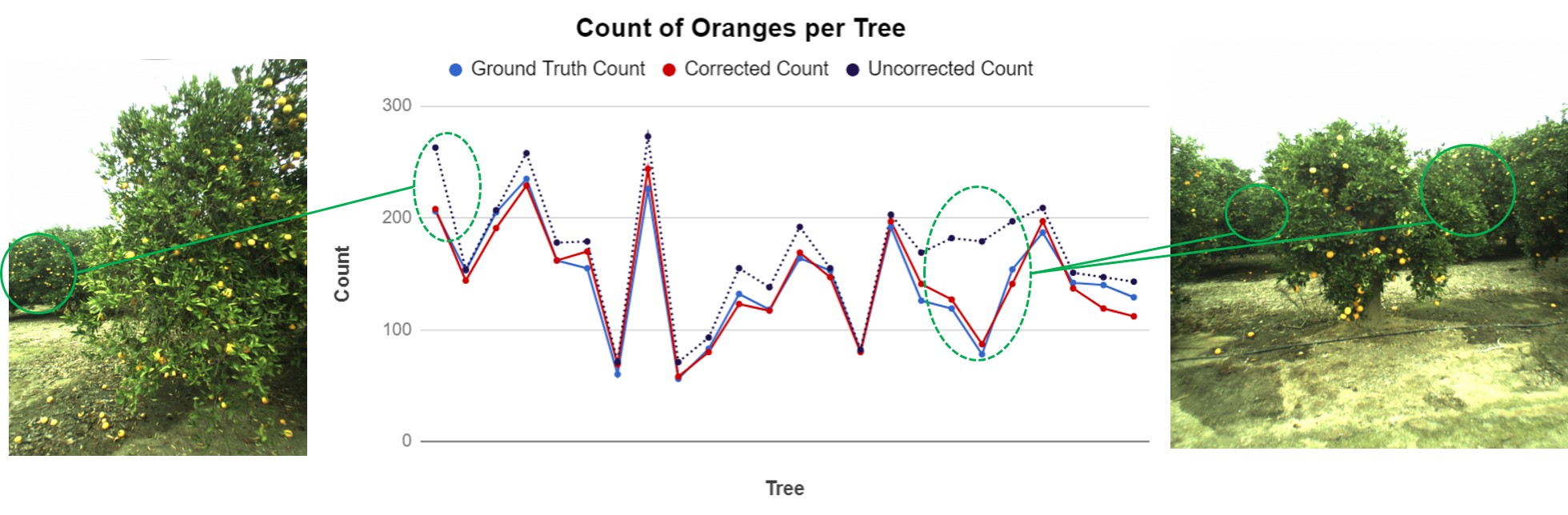}
\caption{\textit{Count of oranges per tree from the corrected model}. The algorithm is able to accurately count the number of oranges in each image sequence. The two green circles in the count chart highlight the two trees in the left and the right images. Our correction significantly improves the count for these two trees since the uncorrected detection and tracking algorithm picks up visible oranges in the background.}
\label{fig:orange_counts}
\vspace*{-0.15in}
\end{figure*}

% \begin{wrapfigure}{R}{0.3\textwidth}
% \centering
% \includegraphics[width=2in]{imgs/size_estimation_oranges.png}
% \caption{\textit{Relative size distribution and outliers rejection for oranges}. X axis is the fruit size divided by mean size of all fruits, Y axis is count for corresponding size. The size distribution is concentrated. The blue line and red line are thresholds for size outliers rejection. All fruits with relative size out of 0.5 - 4 range are rejected.}
% \label{fig:rel_size_orange}
% \vspace*{-0.2in}
% \end{wrapfigure}

% \begin{figure}[t!]
% \centering
% \includegraphics[width=200px]{imgs/plot_major_after_cali_and_size_outliers.png}
% \caption{\textit{Relative Size of oranges after outliers are removed}\todoJD{We can show Fig.~\ref{fig:rel_size_orange} and Fig.~\ref{fig:rel_size_orange_after_outliers} overlaid on the same figure perhaps? Also, how do we know which small fruit detections are because these were occluded fruits?}}
% \label{fig:rel_size_orange_after_outliers}
% \vspace*{-0.2in}
% \end{figure}

In this section, we compare the results of applying our counting pipeline on two different data sets versus human-annotated ground truth counts. We present the count accuracy metrics before and after correction to quantify the benefits of the 3D reconstruction step. These results demonstrate the accurate performance of our algorithm, especially after applying the correction step.

We collected the orange fruit data set during the day with no artificial lighting at Booth Ranch LLC in California. The orange trees were in a non-trellis arrangement. We acquire images of size $1280\times 960$ using a Bluefox USB 2 camera at 10Hz. We collected the apple fruit data set at night using an external flash setup at Washington State. The apple trees were in a trellis arrangement. We acquired images of size $1920\times 1200$ using a PointGrey USB 3 camera at 6Hz.  The orange images were collected with our sensor package~\cite{das2015devices} mounted on a steady cam and carried by human operator at walking speed. The apple images were collected using a utility vehicle driving down the row at around $1~m/s$. The reason for our choice in the two data sets is that they differ in illumination condition, stability of the camera, and fruit occlusion. As a result, they represent different types of data conditions and test the generalizability of our algorithm. 

\begin{figure}[t!]
\vspace*{-0.1in}
\centering
\includegraphics[width=3.7in]{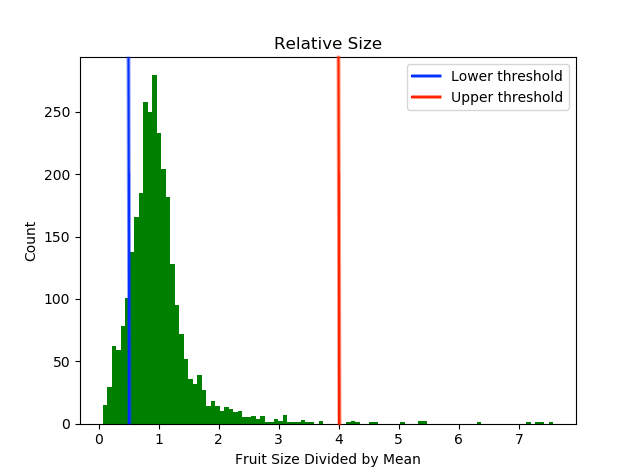}
\caption{\textit{Relative size distribution and outliers rejection for oranges}. The X-axis is the fruit size divided by the mean size of all fruits, and the Y-axis is the count for the corresponding size. The size distribution is concentrated around $1$. The blue line and red line are thresholds for size outliers rejection, where all fruits with a relative size out of the 0.5 - 4 range are rejected.}
\label{fig:rel_size_orange}
\vspace*{-0.25in}
\end{figure}

Our FCN is implemented in Tensorflow. We use the COLMAP package~\cite{schoenberger2016sfm} to generate SfM reconstruction with our preprocessed images and specified feature matching scheme. To measure the performance of our counting pipeline, as well as the effectiveness of our localization and size estimation based count correction, we labeled and counted visual fruits in our two datasets. Based on the ground truth labels and our algorithm outputs, we calculate the count accuracy and standard deviation before and after correction for every tree in our orange dataset. We calculate the similar metrics for every row (approximately 100 frames) in our apple data set since the trees were in a trellis arrangement and were difficult to differentiate.

Fig.~\ref{fig:oranges_count_accuracy} displays the count error metrics for both the orange and apple datasets. For the orange dataset, the uncorrected model has an $L1$ loss of $593$, an error mean of $17.1\%$ and a standard deviation of $26.3\%$. After the 3D reconstruction correction step, the model has an $L1$ loss of $203$, an error mean of $-0.2\%$ and a standard deviation of $7.8\%$, demonstrating a large performance gain in both accuracy and consistency. Fig.~\ref{fig:orange_counts} depicts the various counts for each orange tree for the corrected model, and illustrates the algorithm is obtaining close counts per tree, and not just an accurate average count.

Fig.~\ref{fig:orange_counts} also qualitatively highlights how the 3D reconstruction step corrects for over-counted fruit by removing visible oranges that are in the background. Displayed on the left and right of the chart are the corresponding images of two trees whose uncorrected estimates are inconsistent with the ground truth. In both cases, the trees in another row are visible, a common situation in unstructured farm environments, thus causing our uncorrected counting pipeline to over count. After the correction step, the number of oranges over counted for these two trees is reduced from $57$ to $2$ (ground truth $206$), and $101$ to $9$ (ground truth $78$) respectively. These situations are difficult to identify when only using 2D images, but become easy once the fruit are localized in 3D. 

Fig.~\ref{fig:rel_size_orange} demonstrates that the size distribution of fruits is concentrated, which is expected. Note that these are not sizes in terms of 2D pixel areas, but relative sizes in 3D. The blue and red vertical lines are the lower threshold (0.5) and upper threshold (4) respectively, which are used to reject size outliers and are chosen heuristically. We also set the threshold for depth outliers (fruits from another row of trees) to be 1.5 times the mean depth of all fruits.

\begin{figure}[t!]
\vspace*{0.07in}
\centering
\includegraphics[width=3.4in]{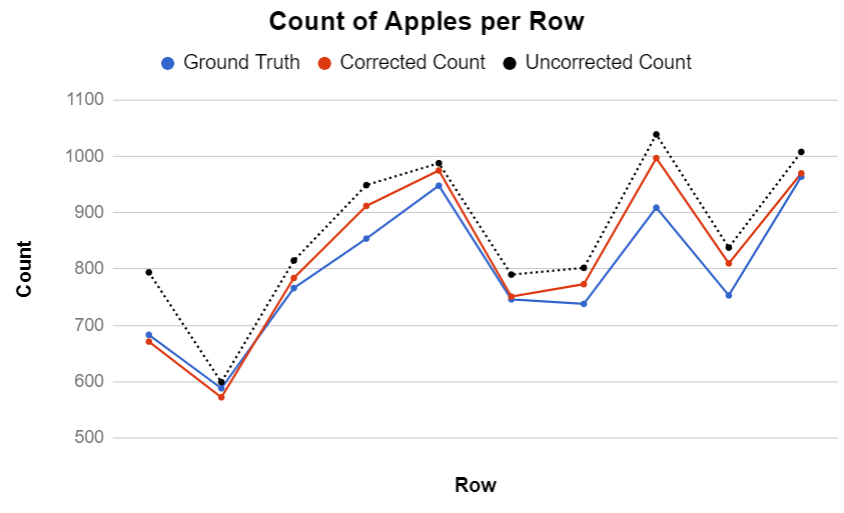}
\caption{\textit{Count of apples per row (a row contains approximately 100 frame images)}. While the uncorrected count performs well due to the controlled environment, the 3D reconstruction correction still improves performance. However, the effect of this correction is much more muted, due to the fact that the structured environment for the apple dataset makes the tracking problem easier.}
\label{fig:apple_counts}
\vspace*{-0.25in}
\end{figure}

For the apple dataset, the uncorrected model has an $L1$ error of $673$, an error mean of $8.5\%$ and standard deviation of $4.7\%$. After the correction step, the model has an $L1$ error is $322$, an error mean of $3.3\%$ and a standard deviation of $4.1\%$. Fig~\ref{fig:apple_counts} displays the various counts for each 100 frames of the apple dataset. While the standard deviation and accuracy of original estimation are already quite good, the 3D correction still further improve results. 

The effect of the 3D reconstruction step on the orange dataset is more significant compared with the apple dataset. The apple dataset is cleaner since the apples are arranged on a trellis resulting in less occlusion, the illumination is artificially controlled at night, and the trees are arranged so that there is less variation in the fruits' distance to camera and less displacement between trees. 

On the other hand, the orange trees are not in a trellis arrangement, so there is much more occlusion and variation in distance. In addition, the videos are taking during the day with natural lighting, so there are more shadows and variations in illumination. Finally, the distance between neighboring trees varies, so many trees in background rows are visible from the camera. As a result, there many more outliers and double counted fruit for oranges, and the effect of the 3D reconstruction step is much more pronounced. In many real world orchards, we would expect an unstructured environment where growers will want to use the algorithm without having to specifically control the video settings. Having an algorithm that can handle these variabilities makes it more widely practical for real-world scenarios.

%As can be seen from Fig.~\ref{fig:tracking_result}, most fruits are recorded as counted fruits before going out of view. In addition, most false positive detections are categorized as new detections or tracked fruits, which will not be added to the total count as detailed above, thus having no negative influence. Our tracking algorithm can work stably over hundreds of frames.

% \subsection{Localization and Size Estimation Results (Maybe do not need this anymore?)}
% Through visually comparison, most of the reconstructed fruit locations are consistent with actual fruit positions in our video frames.
% As for the size estimation, we can see that most fruit sizes distribute around the mean size. This is compatible with the actual distribution of fruit sizes. Using this fruit size distribution, we can easily find false positive fruits – those with too small or too large relative sizes.

\section{Conclusion and Future Work}
We have presented a pipeline consisting of segmentation, tracking, and localization that accurately counts visible fruits across image sequences. To the best of our knowledge, this is the first application of a deep neural network in an algorithm that counts fruit across an image sequence. We evaluated its performance on both an orange data set which features high level of occlusions, depth variation, and uncontrolled illumination, as well as an apple data set which features high color similarity between fruit and foliage. 

In the first stage of our pipeline, we use an FCN to obtain accurate fruit segmentations. We then used the Hungarian Algorithm with the Kalman Filter corrected KLT tracker to track the fruit across image frames. Finally, we localize the fruit in 3D using SfM in order to correct the count by rejecting false positives. We reported an $L1$ error of $203$, error mean of $-0.2\%$ and an error standard deviation of $7.8\%$ for the orange dataset, and and $L1$ error of $322$, error mean of $3.3\%$, and standard deviation of $4.1\%$ for the apple dataset. Our 3D reconstruction correction step improves orange count more significantly compared to the apple dataset since the orange dataset was collected under natural illumination featuring high occlusion and depth variation. Most orchards and farms feature similar uncontrolled conditions, indicating the benefits of our algorithm in practical use scenarios. 

%We also observation is that tracking only the features located near the fruit improved both the speed and quality of the 3D reconstruction. This idea of only tracking features near objects of interest is similar to semantic Simultaneous Localization and Mapping (SLAM)~\cite{Civera2011,Sunderhauf2017} in that we are using features with some semantic meaning, as opposed purely geometric features such corners or lines. When we only care about the location and size of the fruit and a dense reconstruction is not needed, this approach is a promising direction to explore for future work.

One direction of future work is to obtain absolute depth and size. A natural way to obtain these absolute measurements is by fusing IMU measurements. However, another interesting approach is to estimate the average size of the fruit in the physical world, and use this size information to convert the relative size and depth into absolute size and depths while still only using a monocular camera. This approach would be useful when an IMU is not available, which can occur for some types of commodity cameras. 

While we have demonstrated the accuracy of our pipeline on apples and oranges, a second direction of future work would be to evaluate the performance on a variety of other fruits or other objects in the farm. Demonstrating the general counting performance of our algorithm in unstructured environments using only monocular camera images would allow us to explore real-world applications of our algorithms in farms, especially low-resource settings, around the world.

\section{Acknowledgements}
\label{sec:acknowledgements}
This work was supported by USDA NIFA grant 2015-67021-23857 under the National Robotics Initiative. We gratefully acknowledge Anurag Makineni, Andrew Block, and members of the GRASP Laboratory at the University of Pennsylvania for their help with data-collection and annotations. We acknowledge Jim Meyers, and the Washington State Tree Fruit Research Commission for supporting data collection for apples, and Booth Ranches, California for oranges.
% , Yuhpyng Chen, Tien-Chay Chen, Delaney Kaufman, Daniel Orol, DaVonne Henry, Stephanie Kemna, and  others who helped in the labeling effort. We acknowledge the help of Anurag Makineni and Andrew Block in the data collection effort. We dedicate this work to the memory of Stephen Kyle Wilshusen. 

\bibliographystyle{IEEEtran}
\bibliography{ref.bib}

\end{document}